\documentclass[runningheads]{template/llncs}
\usepackage{amssymb, amsmath}
\usepackage{booktabs}
\usepackage{multirow}
\usepackage{graphicx}
\usepackage[acronym]{glossaries}
\usepackage{hyperref}
\usepackage{cleveref}
\usepackage{diagbox}
\usepackage{caption}
\usepackage{adjustbox}
\usepackage{romannum}

\AtBeginDocument{\pagenumbering{arabic}}
\newacronym{ai}{AI}{Artificial Intelligence}
\newacronym{ntd}{NTD}{Neglected Tropical Disease}
\newacronym{fst}{FST}{Fitzpatrick skin type}
\newacronym{ad}{AD}{Atopic dematitis}
\newacronym{eas}{EAS}{Eastern African Set}

\usepackage{color}

\begin{document}
\frenchspacing
\title{PASSION for Dermatology: Bridging the Diversity Gap with Pigmented Skin Images from Sub-Saharan Africa}
\titlerunning{PASSION for Dermatology}
%
\author{
{Philippe Gottfrois}\inst{7, 8} \and 
{Fabian Gröger} \inst{5, 8} \and
{Faly Herizo Andriambololoniaina} \inst{4} \and 
{Ludovic Amruthalingam} \inst{5} \and
{Alvaro Gonzalez-Jimenez} \inst{8} \and 
{Christophe Hsu}\inst{7} \and
{Agnes Kessy} \inst{6} \and
{Simone Lionetti} \inst{5} \and 
{Daudi Mavura} \inst{6} \and 
{Wingston Ng'ambi}\inst{3} \and 
{Dingase Faith Ngongonda} \inst{1} \and
{Marc Pouly} \inst{5} \and 
{Mendrika Fifaliana Rakotoarisaona} \inst{4} \and  
{Fahafahantsoa Rapelanoro Rabenja} \inst{4} \and 
{Ibrahima Traoré} \inst{2} \and
{Alexander A. Navarini}\inst{7, 8}
}


%
\authorrunning{P. Gottfrois et al.}

%
\institute{
Bwaila Hospital Department of Dermatology \and
Clinique dermatologique, Conakry  \and
: Health Economics and Policy Unit, Department of Health Systems and Policy, Kamuzu University of Health Sciences, Lilongwe, Malawi \and
Laboratoire d’Accueil et de Recherche en santé publique spécialisé en TIC \and  
Lucerne University of Applied Sciences and Arts \and
Regional Dermatology Training Center, Moshi \and
University Hospital Basel \and
University of Basel
}
\maketitle              
\begin{abstract} 
Africa faces a huge shortage of dermatologists, with less than one per million people. 
This is in stark contrast to the high demand for dermatologic care, with 80\% of the paediatric population suffering from largely untreated skin conditions. 
The integration of AI into healthcare sparks significant hope for treatment accessibility, especially through the development of AI-supported teledermatology. 
Current AI models are predominantly trained on white-skinned patients and do not generalize well enough to pigmented patients. 
The PASSION project aims to address this issue by collecting images of skin diseases in Sub-Saharan countries with the aim of open-sourcing this data. 
This dataset is the first of its kind, consisting of 1,653 patients for a total of 4,901 images. 
The images are representative of telemedicine settings and encompass the most common paediatric conditions: eczema, fungals, scabies, and impetigo. 
We also provide a baseline machine learning model trained on the dataset and a detailed performance analysis for the subpopulations represented in the dataset.
The project website can be found at \href{https://passionderm.github.io/}{https://passionderm.github.io/}.

\keywords{Dermatology \and Telemedicine Dataset \and Sub-Saharan Africa.}
\end{abstract}
%
%
%
\section{Introduction}
Access to adequate health care remains a global challenge, largely due to a shortage of health workers \cite{naicker2009shortage}. 
This is particularly acute in Africa, which is the continent with the lowest number of health workers per patient, with a shortage of up to 2.4 million doctors and nurses \cite{naicker2009shortage}.
In addition, Africa accounts for 25\% of the global demand for health workers, while having only 1.3\% of the workforce~\cite{naicker2009shortage}.
This phenomenon spans across all medical specialities.
In the case of dermatology, many African countries have less than one dermatologist per million patients \cite{mosam2021dermatology}, making skincare inaccessible to the majority of people.
However, the prevalence of skin conditions can be as high as 87\% in the paediatric population \cite{kiprono2015skin}, a situation that naturally extends to the adult population. 
The consequences of skin diseases are far-reaching and can be physical, psychological and socio-economic, especially in the case of chronic diseases \cite{yew2020psychosocial}.
Teledermatology is a solution that has already been attempted in sub-Saharan Africa \cite{weinberg2009african} to address the lack of access to healthcare. 
However, it does not increase the number of specialists able to provide primary care. 

\looseness=-1
\Gls{ai} holds great promise for scalable diagnosis support, and facilitating access to proper medication and treatment.
Although the field of medical imaging has made significant advances in recent years \cite{pham2021}, a major shortcoming is that publicly available dermatological datasets are significantly biased towards lighter skin tones \cite{daneshjou2022disparities,charrow2017diversity}. 
This bias persists even in private databases, which primarily include local patients, while images of individuals from diverse geographical origins are conspicuously absent \cite{xie2019xiangyaderm,charrow2017diversity}. 
Such a pronounced skew in pigmentation levels does not accurately reflect the diversity of the global population, particularly those without access to traditional dermatologic care. 
Moreover, common ailments in Africa —namely eczematous dermatitis, fungal infections, bacterial infections, and scabies \cite{ranaivo2021clinicoepidemiology,kiprono2015skin}— differ significantly in prevalence from those in Europe \cite{richard2022prevalence} and North America \cite{laughter2020burden}, which are the usual sources of patient data for public datasets.
Efforts were made to address this disparity, but existing datasets are often small \cite{daneshjou2022disparities} or fail to capture the spectrum of prevalent skin conditions in Africa, particularly \glspl{ntd} \cite{daneshjou2022disparities}.


This paper aims to fill a critical gap in dermatological research by collecting a comprehensive and inclusive dataset from regions where access to dermatologists is difficult. 
To achieve this goal, a prospective collection of dermatological cases in sub-Saharan Africa was initiated in 2020.
Several countries have collaborated in a global effort to create this dataset, which is intended to serve not only as a valuable resource for training machine learning algorithms and evaluating existing models in these underrepresented populations, but also as a resource for training medical personnel and helping to enable \gls{ai}-driven teledermatology systems. 
The dataset consists of patient images, with phototypes \Romannum{4}, \Romannum{5} and \Romannum{6} on the \gls{fst} scale, acquired in conditions similar to teledermatology consultation, and representative of the most common skin conditions in sub-Saharan Africa.
These conditions cover up to 80\% of the conditions seen in the paediatric population \cite{10.1001/archderm.134.11.1363}. 
During the data collection process, substantial effort was devoted to including the paediatric population, as this is the demographic group most affected by skin disease.
This results in a dataset that represents the diversity observed in these countries and effectively covers the dermatological conditions of children. 
Additionally, we provide a baseline machine learning model with a corresponding in-depth performance analysis for subpopulations, showcasing evaluation possibilities with this dataset. 
By releasing the PASSION dataset, we aim to contribute to the advancement of dermatological research and the equity of healthcare solutions.

\section{Related Work} 
Existing datasets predominantly contain cases with lighter skin tones from North American and European populations \cite{daneshjou2022disparities}.
In addition, only a minority of datasets include skin type labels, such as the \gls{fst} scale, or other information related to ethnicity \cite{kawahara2018seven,tschandl2018ham10000,combalia2019bcn20000,rotemberg2021patient,giotis2015med,mendoncca2013ph,sun2016benchmark}.
For example, the Fitzpatrick17k dataset \cite{groh_evaluating_2021} contains 16,577 clinical images annotated with \gls{fst} labels.
However, concerns were raised about the quality of the dataset as it was found to contain many duplicates, irrelevant samples and label errors \cite{groger_selfclean_2023}.
Despite its emphasis on skin type labels, it contains predominantly lighter skin types, illustrating the challenges of achieving true diversity.
On the other hand, the Diverse Dermatology Images (DDI) dataset \cite{daneshjou2022disparities} explicitly focuses on different skin tones and contains 656 clinical images evenly distributed across \gls{fst} types.

Other work has explored the use of generative networks \cite{akrout2023diffusionbased,sagers2022improving} to generate synthetic images of phototypes \Romannum{4}, \Romannum{5} and \Romannum{6}, aiming to increase the diversity of dermatological datasets and improve the robustness of classifiers. 
However, these efforts face inherent limitations since they may not fully capture clinical realism, potentially introducing inaccuracies.
For example, a generative model might produce an image of a melanoma on highly pigmented skin. 
As a sign of insufficient realism, the image might contain easily visible erythema and a hypopigmented border that does not occur in real life. 
Conversely, an image of acne on lightly pigmented skin could contain residual hyperpigmentation when in reality this occurs almost exclusively on darker skin types.

The spectrum of skin conditions covered by dermatology datasets illustrates another critical facet of diversity. 
The majority of existing resources focus narrowly on pigmented lesions, leaving a vast array of non-pigmented skin diseases and rare conditions underrepresented \cite{kawahara2018seven,tschandl2018ham10000,combalia2019bcn20000,rotemberg2021patient,giotis2015med,mendoncca2013ph,sun2016benchmark}. 
In addition, most of the images are dermoscopy images, which require a specific tool and some expertise to capture. Such images tend to be highly standardised. 
This not only limits the scope of research, but also the applicability of the resulting diagnostic models.
Initiatives such as Fitzpatrick17k~\cite{groh_evaluating_2021}, DDI~\cite{daneshjou2022disparities}, and SKINCON~\cite{daneshjou2023skincon} are the only ones encompassing a wider range of skin conditions in European and North American populations, both non-pigmented and rare, in the form of non-standardised clinical images.

\section{The PASSION dataset} 
\label{sec:PASSION}

\begin{figure}[t]
  \begin{minipage}[b]{0.49\textwidth}
    \centering
    \includegraphics[width=\textwidth]{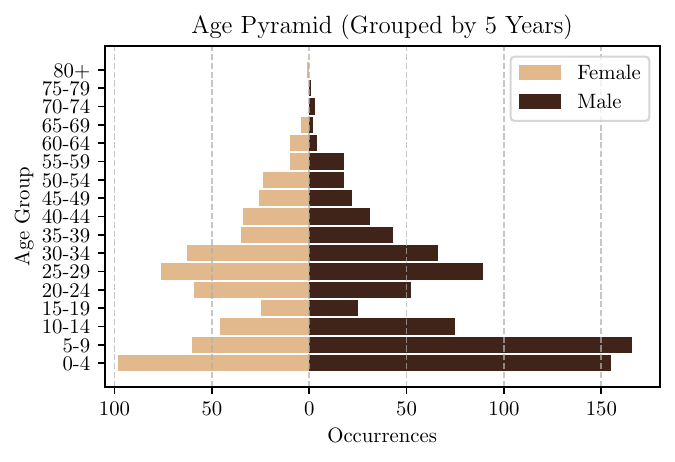}
    \captionof{figure}{Age distribution per gender.}
    \label{fig:age_pyramid}
  \end{minipage}
  \begin{minipage}[b]{0.49\textwidth}
    \centering
\includegraphics[width=\textwidth]{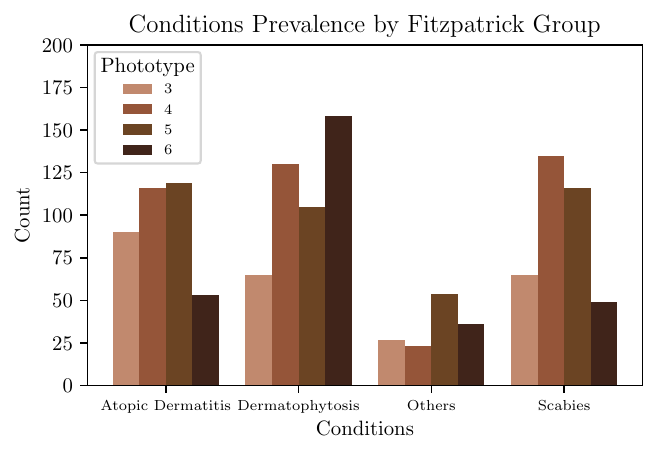}
    \captionof{figure}{Conditions per skin phototypes.}
    \label{fig:conditions_fitz}
  \end{minipage}
\end{figure}

\subsubsection{Data collection.}
Data were prospectively collected from 2020 to 2023 in Madagascar, Guinea, Malawi, and Tanzania. Madagascar's cases (985 cases, 2,185 images) were nationwide, Guinean cases (384 cases, 480 images) were in Conakry, Malawi's cases (261 cases, 2,150 images) were in Lilongwe, and Tanzanian cases (23 case, 86 images) were in Moshi. Malawi and Tanzania data were aggregated as the Eastern African Set (EAS) due to dataset size and geographical proximity.
Local board-certified dermatologists acquired the collections, annotated and uploaded the data in each country, either individually or in teams. 
In addition to images and diagnoses, clinically relevant information such as age, sex and body localisation was collected. 
Diagnoses were converted to ICD-10 and ICD-11 codes \cite{icd11}, the latest two WHO standards for disease classification, to facilitate integration and interoperability with existing and future datasets.
The data collection process reflects a teledermatology setup, as image capture was performed using mobile phones commonly used in the collection areas. 
Multiple images of the same case were taken to present lesions from different angles to ensure a comprehensive and detailed presentation and to mitigate the risks associated with low-quality images.
Diagnoses were made by clinical examination and patient follow-up.  

\subsubsection{Ethical approval.}
Data collection was performed with patients' informed consent in accordance to the respective legislation and ethics guidelines of each contributing country. Analysis of anonymized images was performed according to Swiss ethics statement EKNZ-2018-01074.

\subsubsection{Patient privacy.}
After the collection phase, manual anonymization was conducted to ensure individuals' privacy, recognizing that clinical dermatology images can include identifiable features. This involved removing identifiable features from the images by inpainting or cropping the images. In addition, depictions of non-disease-involved private body locations were minimized as much as possible.

\begin{table}[t]
    \centering
    \caption{Number of cases per country and body localization.}
    \label{tab:Distribution_Cases}
    \setlength{\tabcolsep}{2pt}
    \resizebox{0.9\linewidth}{!}{%
    \begin{tabular}{l|*{3}{r}|*{9}{r}}
      \toprule
        & \multicolumn{3}{c|}{\textbf{Country}}
        & \multicolumn{9}{c}{\textbf{Body Localization}} \\
      \textbf{Conditions}
        & Guinea & Madagascar & \acrshort{eas}
        & Arm & Back & Pelvic & Face & Foot & Hair & Hand & Leg & Torso \\
      \midrule
      Eczema 
        & 57 & 301 & 56 
        & 41 & 38 & 11 & 76 & 37 & 24 & 23 & 29 & 72 \\
      Fungal 
        & 230 & 299 & 50
        & 32 & 33 & 23 & 33 & 35 & 198 & 36 & 41 & 109 \\
      Scabies 
        & 58 & 356 & 57
        & 13 & 147 & 54 & 1 & 21 & 0 & 76 & 10 & 126 \\
      Others 
        & 39 & 29 & 121
        & 14 & 61 & 8 & 41 & 3 & 13 & 1 & 21 & 16 \\
      \bottomrule
    \end{tabular}
    }
\end{table}

\subsubsection{Data cleaning.}
In order to ensure the quality and accuracy of the dataset, extensive data cleaning was carried out after acquisition.
However, traditional data cleaning is very resource intensive as it usually involves checking each sample or pair of samples.
Therefore, we resort to a more resource-efficient data cleaning protocol consisting of the use of an algorithmic data cleaning strategy followed by manual verification \cite{pmlr-v225-groger23a}.
We checked for irrelevant samples, near duplicates and label errors.
For irrelevant samples and near duplicates, two expert labellers performed manual verification. 
Similarly, verification for label errors was performed by two board-certified dermatologists with experience practising in sub-Saharan Africa. 
All experts involved in data curation were not involved in data collection.
Annotations were aggregated by unanimous agreement.
Verified irrelevant samples were removed, near duplicates were combined by assignment to the same case, and label errors were corrected.

\subsubsection{Distributions.}
Skin phototypes vary between collection centers and are representative of the local population, with a majority of type \Romannum{6} on the \gls{fst} scale in Guinea. 
\acrshort{eas} and Madagascar have a fairly similar distribution from type \Romannum{3} to type \Romannum{5} or \Romannum{6}. 
The age distribution is shown in \cref{fig:age_pyramid}, with the \acrshort{eas} images including only paediatric cases. 
The gender distribution of the collection shows a slight tendency towards male cases: Guinea 62\%, Madagascar 56\% and \acrshort{eas} 65\%.
The largest difference is in the 5-10 year age group, where the number of male cases is 2 to 3 times higher. 
The cases collected represent 44 conditions grouped into 4 categories: Eczema, Fungal, Scabies (a skin \gls{ntd}~\cite{engelman2021framework}) and Others. 
The grouping is based on the ICD-10 ontology. 
In addition, a binary column ``Impetiginized'' is added, as impetigo may occur alone or in combination with other conditions. 
There are a total of 191 (11.5\%) impetiginised cases.  
Recognising its presence is clinically relevant as the prescription of drugs such as corticosteroids for eczema could increase the development of the bacteria by reducing the immune response. 
The prevalence of conditions is shown in \cref{tab:Distribution_Cases}.

\subsubsection{Data split.}
The dataset has a pre-defined 80/20 stratified development-test split to ensure reproducibility and fair comparison of future results. 
The split is performed at the patient level to prevent information leakage leading to overly optimistic results. 

\subsubsection{Availability.}
The dataset can be downloaded at\href{https://passionderm.github.io/}{https://passionderm.github.io/} and is released under is released under the PASSION datasets public license.

\section{Methodology}

\subsubsection{Model architecture.}\label{subsub:Model}
We use transfer learning to train a ResNet-50~\cite{he_deep_2016} pre-trained on ImageNet~\cite{deng_imagenet_2009}.
We replace the last fully connected classification layer with the following sequence of layers: a dropout layer with a 30\% chance of dropping, batch normalisation, and a single linear layer generating the class activations.
The models are trained using Adam optimisation to minimise the weighted cross-entropy loss.
The optimal learning rate was found using the learning rate range test described in \cite{smith_cyclical_2017}.
All models are trained by first unfreezing only the classification head for 10 epochs, followed by unfreezing all layers until the end of training.
Images are normalized using mean and standard deviation of ImageNet~\cite{deng_imagenet_2009}.
To improve the generalisation of our models and to avoid overfitting, we perform different data augmentations: random resizing and cropping the images to $224\times224$ pixels, random horizontal and vertical flipping, and random rotation between 0 and 90 degrees.

\subsubsection{Evaluation protocols.}
For hyperparameter tuning, 5-fold cross-validation is used on the development set, while the test set was kept separate.
We evaluate the classifiers with different schemes in order to provide an in-depth analysis.
\begin{enumerate}
    \item Performance in predicting skin conditions and detecting impetiginised cases on the dedicated hold-out test set.
    \item Generalisation from two centers to the wider population available in the dedicated hold-out test set, which contains samples from both the two known centers and one completely unknown centre.
    \item Generalisation from different age groups (i.e. paediatric and adolescent cases) to the broader population available in the dedicated hold-out test set, which contains samples from the known age group and a completely unseen one.
\end{enumerate}
All models are trained until the validation loss does not improve over twenty consecutive epochs. 
Performance is measured in terms of balanced accuracy and macro-averaged precision and sensitivity for detecting skin conditions, and binary F$_1$, precision, and sensitivity for detecting impetigo.

\subsubsection{Implementation.}
The implementation is based on PyTorch 1.9~\cite{paszke_pytorch_2019}. The experiments are run on an Nvidia DGX station with eight V100 GPUs, each with 32 GB of memory, 512 GB of system memory, and 40 CPU cores. 
Code to replicate our results is available at \href{https://passionderm.github.io/}{https://passionderm.github.io/}.

\section{Results}

\begin{table}[t]
    \centering
    \caption{
        Hold-out test performance of different models for predicting skin conditions (i.e. Eczema, Fungal, Scabies, and Others) or detecting impetiginized cases (i.e. binary classification).
    }
    \label{tab:Experiment1-Results}
    \setlength{\tabcolsep}{4pt}
    \resizebox{0.9\linewidth}{!}{%
    \begin{tabular}{l|*{3}{c}|*{3}{c}}
      \toprule
        & \multicolumn{3}{c|}{\textbf{Conditions}}
        & \multicolumn{3}{c}{\textbf{Impetigo}} \\
      \textbf{Model Name}
        & \multicolumn{1}{c}{bal. Acc.} 
        & \multicolumn{1}{c}{Precision} 
        & \multicolumn{1}{c|}{Sensitivity}
        
        & \multicolumn{1}{c}{F$_1$} 
        & \multicolumn{1}{c}{Precision} 
        & \multicolumn{1}{c}{Sensitivity} \\
      \midrule
         Random Prediction 
            & 0.25 & 0.25 & 0.25
            & 0.27 & 0.18 & 0.56 \\
         Constant Prediction
            & 0.25 & 0.06 & 0.25
            & 0.28 & 0.16 & 1.00 \\
         \midrule
         Fine-tuned ImageNet 
            & 0.70 & 0.70 & 0.70
            & 0.63 & 0.52 & 0.79 \\
      \bottomrule
    \end{tabular}
    }
\end{table}

In \cref{tab:Experiment1-Results}, we compare the results of a classifier with two uninformed baselines, i.e. random prediction and constant class prediction (the majority class for detecting conditions and the positive class for detecting impetigo).
For skin condition prediction, the fine-tuned ImageNet model achieved a balanced accuracy score of 0.70, a precision of 0.70 and a sensitivity of 0.70, significantly outperforming both uninformed baselines.
In the specific task of impetigo detection, the fine-tuned ImageNet model achieved a binary F$_1$ score of 0.63, with precision and sensitivity rates of 0.52 and 0.79, respectively.
Models were also compared at subject level by aggregating the prediction for each image of the subject using majority voting.
This evaluation is more similar to the task of teledermatology.
Here, the fine-tuned ImageNet achieves a balanced accuracy of 0.64 for skin disease prediction and binary F$_1$ of 0.47 for impetigo detection, which is 6\% and 16\% lower, respectively, than in the sample-level evaluation.
This difference highlights the importance of evaluating at both sample and subject-level to get a more complete picture of a classifier's performance.
In \cref{tab:Experiment1-Demography}, we detail the performance of the classifier for each skin type (i.e. \gls{fst} \Romannum{3}-\Romannum{6}) and gender present in the dataset.
The results show that the model has small performance variations across skin types, with a performance gap between the best and worst performing class of 7\% (\gls{fst} \Romannum{4} vs. \Romannum{6}) and 11\% (\gls{fst} \Romannum{4} vs. \Romannum{3}) in terms of balanced accuracy for predicting skin conditions and binary F$_1$ for detecting impetigo, respectively.
The performance for female and male cases is only slightly different for both prediction tasks.

\begin{table}[t]
    \centering
    \caption{
        Detailed hold-out test performance for predicting skin conditions or detecting impetiginized cases for a fine-tuned ImageNet model.
        Performance is shown for each skin type and gender in the dataset separately.
    }
    \label{tab:Experiment1-Demography}
    \setlength{\tabcolsep}{4pt}
    \resizebox{0.9\linewidth}{!}{%
    \begin{tabular}{l|l|*{3}{r}|*{3}{r}}
      \toprule
        &
        & \multicolumn{3}{c|}{\textbf{Conditions}}
        & \multicolumn{3}{c}{\textbf{Impetigo}} \\
      \textbf{Demography}
        & \multicolumn{1}{c|}{\textbf{Support}} 
        
        & \multicolumn{1}{c}{bal. Acc.}
        & \multicolumn{1}{c}{Precision} 
        & \multicolumn{1}{c|}{Sensitivity}
        
        & \multicolumn{1}{c}{F$_1$} 
        & \multicolumn{1}{c}{Precision} 
        & \multicolumn{1}{c}{Sensitivity} \\
      \midrule
        \gls{fst} \Romannum{3} & 303
            & 0.72 & 0.71 & 0.72
            & 0.68 & 0.61 & 0.77 \\
        \gls{fst} \Romannum{4} & 254
            & 0.66 & 0.66 & 0.66
            & 0.57 & 0.47 & 0.72 \\
        \gls{fst} \Romannum{5} & 341
            & 0.70 & 0.69 & 0.70
            & 0.62 & 0.49 & 0.85 \\
        \gls{fst} \Romannum{6} & 88
            & 0.73 & 0.74 & 0.73
            & 0.62 & 0.52 & 0.79 \\
        \midrule
        Female & 425
            & 0.72 & 0.72 & 0.72
            & 0.64 & 0.52 & 0.82 \\
        Male & 561
            & 0.68 & 0.69 & 0.68
            & 0.61 & 0.52 & 0.76 \\
      \bottomrule
    \end{tabular}
    }
\end{table}

To investigate whether a model is able to generalise across collection centers, we compare the performance of a model trained only on the subset of samples from two collection centers with a model trained on all three collection centers on the dedicated test set.
To ensure an equal number of samples between the subsets, we randomly resample the training dataset to the same size.
Models trained on only two centers achieve a balanced accuracy of 0.61 $\pm$ 0.05, while models trained on all centers achieve 0.68 $\pm$ 0.01.
The lowest generalisation was obtained by training only on cases from Guinea and Madagascar, probably due to the predominance of paediatric cases in \acrshort{eas}. 
The difference in performance between the two groups highlights the problem of generalising across collection centers and also highlights the importance of collecting data from different sources.

Similarly to the above, we investigate whether a model is able to generalise across age groups, i.e. between paediatric and adolescent cases.
Here we compare the performance of a model trained only on paediatric images with a model trained on both paediatric and adolescent cases and vice versa.
Resampling is used to ensure an equal number of samples within the groups.
Models trained only on paediatric images (i.e. age $\leq 16$) achieve a balanced accuracy of 0.66, while models trained on all age groups achieve a score of 0.70, indicating possible generalisation from paediatric images to the wider population.
On the other hand, models trained only on adolescent images (i.e. age $> 16$) achieve a balanced accuracy of 0.43, while models trained on all age groups achieve a score of 0.68.
This indicates that generalisation from paediatric cases is higher than from adolescent cases, further highlighting the importance of focusing on the collection of paediatric datasets such as PASSION.

\section{Conclusion}
This paper presents the PASSION dataset, a pioneering step towards bridging the significant diversity gap in dermatological datasets, specifically targeting pigmented skin types prevalent in sub-Saharan Africa. 
Our efforts culminate in the compilation of a comprehensive dataset that not only improves the representation of darker skin tones in dermatological studies but also facilitates the development of AI-driven diagnostic tools that are inclusive. 
In addition, we provide an in-depth analysis of a baseline machine learning model trained and evaluated on PASSION.
Future work will focus on expanding the dataset to include a wider range of skin conditions and exploring innovative machine learning algorithms that can further adapt to the challenge of diagnosing skin diseases across a diverse range of skin types.

\begin{credits}
\subsubsection{\ackname} This study was funded by grant REG-19-024 awarded by Fondation Botnar, Basel, Switzerland.

\subsubsection{\discintname}
The authors have no competing interests to declare that are relevant to the content of this article.
\end{credits}

%
\bibliographystyle{template/splncs04}
\bibliography{bibliography}

\end{document}